\documentclass[pdflatex,sn-mathphys]{sn-jnl}
\jyear{2021}%
\theoremstyle{thmstyleone}%
\theoremstyle{thmstyletwo}%
\newtheorem{remark}{Remark}%
\newtheorem{pref}{Preference function}
\theoremstyle{thmstylethree}%
\newtheorem{definition}{Definition}%
\newtheorem{lemma}{Lemma}

\DeclareMathOperator*{\argmin}{arg\,min}
\DeclareMathOperator*{\sequ}{seq}
\DeclareMathOperator*{\rank}{rank}
\DeclareMathOperator*{\crowd}{crowd}

\newcommand{\Rr}{\mathbb{R}}
\newcommand{\Pp}{\mathcal{P}}
\newcommand{\F}{\mathcal{F}}
\newcommand{\Ss}{\mathcal{S}}

\usepackage[capitalise]{cleveref}

\usepackage{arydshln}
\usepackage{subfigure}

\usepackage{comment}

\usepackage{etoolbox}
    \makeatletter
    \patchcmd{\ps@headings}
    {\hbox to \hsize{\hfill Springer Nature 2021 \LaTeX\ template\hfill}}
    {\hbox to \hsize{\hfill \hfill}}
    {}
    {}
    \patchcmd{\ps@titlepage}
    {\hbox to \hsize{\hfill Springer Nature 2021 \LaTeX\ template\hfill}}
    {\hbox to \hsize{\hfill \hfill}}
    {}
    {}
    \makeatother
\begin{document}

\title[Optimal Neural Architectures]{Automatic selection of the best neural architecture for time series forecasting}

\author[1]{\fnm{Qianying} \sur{Cao}}\email{qianying\_cao@brown.edu}

\author[1]{\fnm{Shanqing} \sur{Liu}}\email{shanqing\_liu@brown.edu}

\author[2]{\fnm{Alan} \sur{John Varghese}}\email{alan\_john\_varghese@brown.edu}

\author[1]{\fnm{Jerome} \sur{Darbon}}\email{jerome\_darbon@brown.edu}

\author[3]{\fnm{Michael} \sur{Triantafyllou}}\email{mistetri@mit.edu}

\author*[1]{\fnm{George} \sur{Em Karniadakis}}\email{george\_karniadakis@brown.edu}

\affil[1]{\orgdiv{Division of Applied Mathematics}, \orgname{Brown University}, \orgaddress{\city{Providence}, \postcode{02906}, \state{Rhode Island}, \country{U.S.A}}}
\affil[2]{\orgdiv{School of Engineering}, \orgname{Brown University}, \orgaddress{\city{Providence}, \postcode{02906}, \state{Rhode Island}, \country{U.S.A}}}
\affil[3]{\orgdiv{Department of Mechanical Engineering}, \orgname{Massachusetts Institute of Technology}, \orgaddress{ \city{Cambridge}, \postcode{02139}, \state{Massachusetts}, \country{U.S.A}}}

\abstract{
Time series forecasting is essential across domains such as healthcare, energy, and climate modeling. While models like LSTMs, GRUs, Transformers, and State-Space Models (SSMs) have become widely used, selecting the optimal architecture remains unclear. We propose an automated framework that systematically designs hybrid architectures by combining LSTM, GRU, attention, and SSM modules. Our approach uses multi-objective optimization to explore combinations and orderings of blocks, yielding Pareto-optimal architectures that balance user-defined trade-offs among objectives. A preference function selects the most suitable model for a given application. Moreover, two sampling-based iterative procedures for Pareto-front exploration are introduced, which reduces the total training cost by nearly eightfold. Across four real-world benchmarks, our framework reveals that simple models excel in speed, while hybrid compositions dominate when balancing accuracy and complexity. Our findings challenge the notion of a universally superior neural architecture, emphasizing instead the value of data- and objective-driven design in time series forecasting.}


\maketitle

\section*{Introduction}\label{sec1}
Time series forecasting plays an important role across diverse fields, including weather and environmental science, healthcare, finance, and engineering systems. Developing an efficient and reliable forecasting model is crucial for applications, such as health monitoring, digital twins, and decision-making in these domains.

A plethora of deep learning approaches have been developed for time series forecasting~\cite{zhang2003time,deb2017review,lim2021time,torres2021deep,cao2024laplace,wang2024timemixer}, with the Long Short-Term Memory (LSTM), Gated Recurrent Unit (GRU), Transformers, and State-Space Models (SSMs) the most widely used neural network models.  LSTM~\cite{hochreiter1997long} was proposed in 1997 to mitigate the vanishing gradient problem in traditional recurrent neural networks, and advanced versions~\cite{graves2005framewise,siami2019performance,choi2018combining} were developed to solve different engineering problems~\cite{tsai2020learning,tang2021short,cao2024deep}. Compared with LSTM, GRU~\cite{cho2020learning}, which has a gating mechanism to input but lacks an output gate, employs fewer parameters. 
The Transformer architecture, which relies solely on Attention mechanisms, has become highly popular for translation tasks since it was first introduced in 2017~\cite{vaswani2017Attention}. The multi-headed self-Attention replaced the recurrent layers, which are typically used in encoder-decoder architectures. Several examples of the application of Transformers are presented in~\cite{chitty2022neural, pokhrel2022transformer,khan2022transformers,islam2023comprehensive,zeng2023transformers,varghese2024transformerg2g,chen2024tempronet,song2024stvformer,wang2024comprehensive}. 
To address the Transformer's computational inefficiency on long sequences, Gu and Dao~\cite{gu2023mamba} proposed Mamba, which integrated selective SSMs into a simple network architecture without Attention, achieving faster inference and better performance than the Transformer models on diverse application domains. Following this, 
several advanced types of Mamba and their applications were rapidly developed~\cite{liang2024bi,li2024cmmamba,wu2024umambatsf,wang2024mamba,hu2024time,hu2024state,cheng2024mamba,hu5149007deepomamba}. 

From the literature, it is evident that LSTM, GRU, Transformer, and Mamba are the most widely used tools for time series forecasting, and each of them exhibits unique characteristics. However, determining the ``best" model remains challenging, as performance comparisons often vary depending on the evaluation metrics and datasets employed. Moreover, hybrid models which combine these architectures have emerged~\cite{zhang2021multiscale,abbasimehr2022improving,stefenon2023wavelet,ma2024fmamba,zhang2024integration,lieber2024jamba,hu5149007deepomamba}, significantly expanding the design space of network architectures. This abundance of available models can overwhelm researchers and engineers, making it challenging to assess their relative performance and choose the most suitable one. 
This challenge is closely related to the field of Neural Architecture Search (NAS), which aims to automatically discover high-performing architectures within a large design space. A rich body of work has explored NAS using reinforcement learning, evolutionary algorithms, and gradient-based or differentiable
methods~\cite{zoph2016neural,pham2018efficient,real2019regularized,liu2019darts,elsken2019neural,chen2019progressive,guo2020single,wei2022npenas,xue2023evolutionary,liu2025neural}. 
These approaches have achieved impressive results in diverse domains. Our work is not intended to compete with global NAS pipelines, but serves as a reliable refinement state that can be applied after candidate architectures are obtained.
However, most NAS methods are primarily tailored to single-objective settings and do not explicitly target the multi-objective, application-driven trade-offs that arise in practical forecasting scenarios.

\begin{figure}[ht]
\centering
\includegraphics[width=\textwidth, trim = 0cm 9.5cm 0cm 0cm, clip]{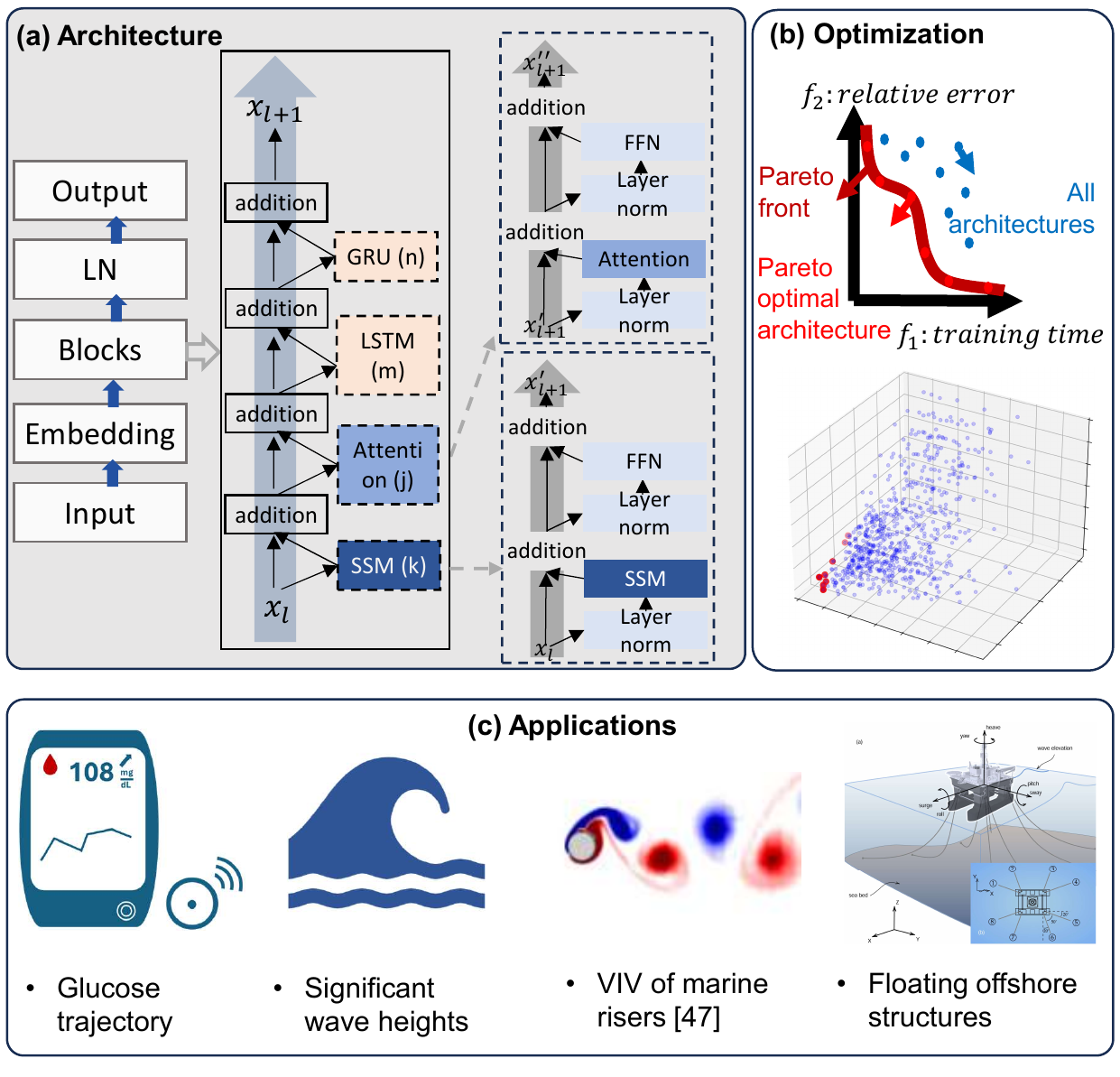}\\
\caption{(a) A schematic representation of the entire parametrized architecture, (b) Multi-objective optimization and Pareto optimality, and (c) four real-world applications studied in this work~\cite{wang2021large}.}\label{Figure_architecture}
\end{figure}

In this paper, we propose an automated general framework designed to identify the best composite architectures, incorporating LSTM, GRU, multi-head Attention, and SSM blocks. The number of each type of blocks and the sequence of these blocks are treated as hyperparameters. 
This flexible framework not only includes many existing models but, more importantly, automatically generates new network models that are not yet found in the literature. 
One such discovered architecture is illustrated in \autoref{Figure_architecture}(a). 
Additionally, each architecture requires the tuning of other hyperparameters, such as the hidden dimension. 

The composite architectures we propose naturally raise the  
question of how to select the optimal design. 
To address this challenge, we propose a multi-objective optimization (MOO) method to automatically select the best network architecture with the best hyperparameters (see \autoref{Figure_architecture}(b)).
In particular, we first parameterize the construction of different architectures within a given domain. 
Next, we model the performance of these architectures using a vector-valued map, which maps a parameterized architecture to objectives that reflect its performance, such as relative $\mathcal{L}_2$ error, training time, and the number of parameters. 
The selection of best architecture is then formulated as a multi-objective optimization problem, in the sense of minimizing this vector-valued map over the parameterization space. 

We applied our methodology to four real-world applications (see a sketch in \autoref{Figure_architecture}(c)). Through computational experiments, we observe that this multi-objective optimization problem does not admit a classical solution that optimizes all performance metrics. 
In other words, no single architecture achieves the best performance across all criteria, including relative $\mathcal{L}_2$ error, training time, and the number of parameters. 
Therefore, we adopt a weaker sense of optimality -- Pareto optimality. In this context, no other architecture can outperform the Pareto optimal ones across all criteria. We then identify all Pareto optimal architectures for different applications. Furthermore, we propose a preference function-based discovery methodology. Using a user-defined preference function, we can discover the architecture on demand. 
Finally, we introduce a Linear Programming (LP) approach to rediscover a specific known architecture from those on the Pareto front. Specifically, for a given Pareto optimal architecture, we identify a preference function under which this architecture is the best. A key practical challenge in this setting is the offline cost of training a large number of candidate architectures, most of which are ultimately non–Pareto-optimal. To reduce this cost, we introduce two sampling-based iterative procedures that explore the Pareto front directly in the parameterized architecture space. While the first method minimizes computational overhead by locally sampling around current Pareto-optimal designs, the second leverages the NSGA-II framework to ensure a comprehensive search across the objective space.

We summarize the main contributions of this work in the following:
\begin{itemize}
\item[i.] We introduce composite architectures, which are hybrids consisting of GRU, LSTM, multi-head Attention, and SSM blocks to leverage the best features of each. By tuning the number and the sequence of each type of blocks, a plurality of architectures can be created based on this composite model. 
\item[ii.] We formulate the problem of selecting the best architecture in a constrained, enumerated design space as a multi-objective optimization problem, where the objective function models the multi-dimensional performance criteria. The set of Pareto optimal architectures is identified, and the best architecture is tailored to specific engineering needs through a preference-based approach. 
\item[iii.] We apply our method to four real-world applications, which demonstrate that the best-performing model emerges as a data-driven composite architecture tailored to user-defined criteria. 
\item[iv.] To reduce offline training cost, we introduce two sampling-based iterative procedures. Empirically, three sampling rounds (about $13.5\%$ of candidates evaluated) already yield a high-quality approximation of the full Pareto front. 
\item[v.] We propose an architecture-transfer strategy that adapts a source optimal architecture to same domain trajectories by lightweight local refinement, which achieves almost two orders of magnitude improvement in search efficiency.
\end{itemize}

\section*{Results}\label{sec2} 
We apply our model to four real-world applications, including biomedicine, weather, and flow-structure interactions, as shown in \autoref{Figure_architecture}(c). For each case, we define the hyperparameter range, including block numbers, block sequence, and hidden dimension. This setup generates multiple architectures, each producing distinct forecasting results. 
We employ the multi-objective optimization methodology to select all Pareto optimal architectures. 
Furthermore, by applying preference functions tailored to engineering needs, 
we identify the best architecture for each specific context. The final part of this section demonstrates the effectiveness of the proposed iterative architecture sampling methods using the glucose prediction task as a representative example.

\subsection*{Predicting glucose trajectory}
\begin{figure}[ht]
\centering
\includegraphics[width=\textwidth, trim = 0cm 13cm 0cm 0cm, clip]{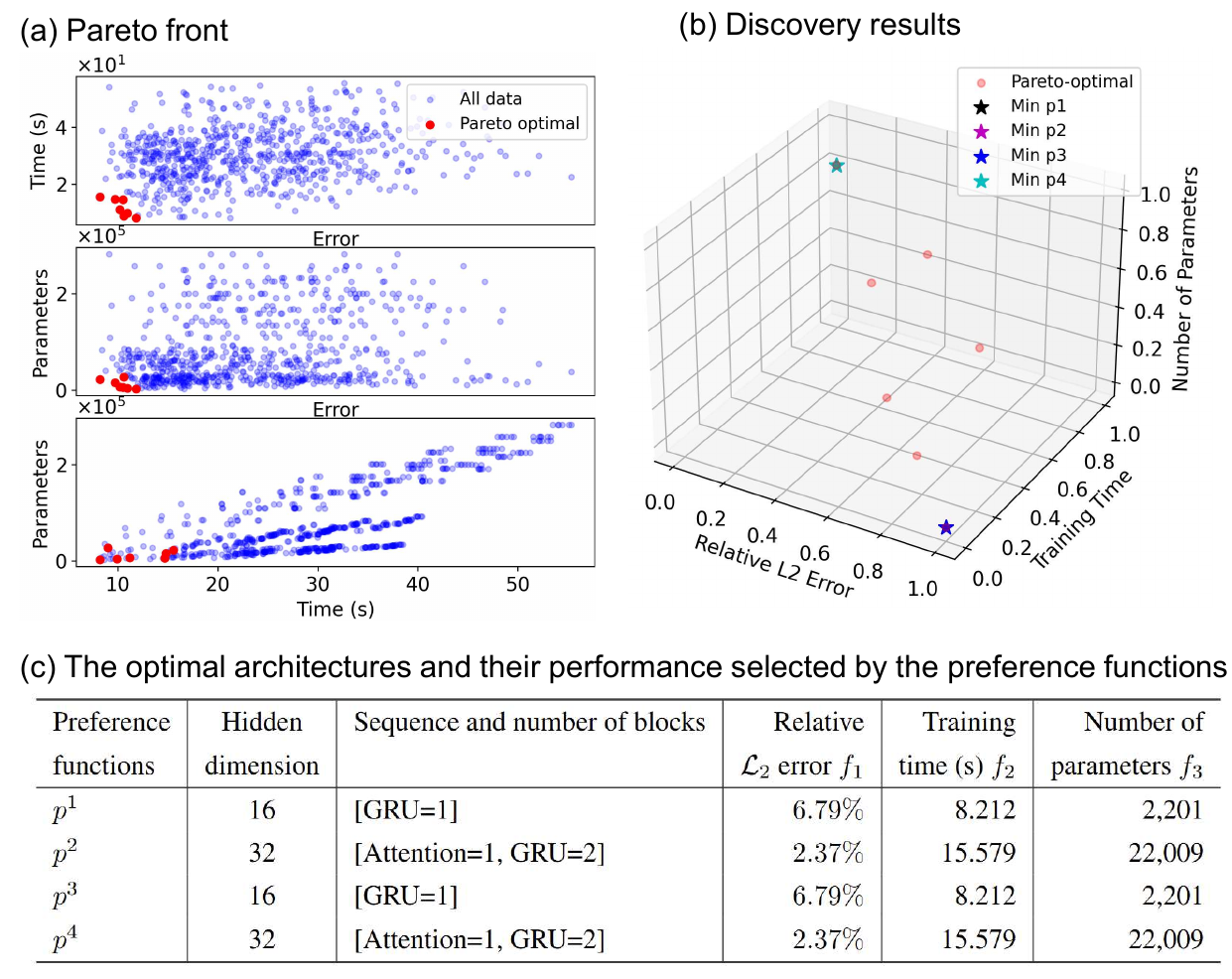}\\
\caption{Glucose prediction: (a) Pareto front showing relative error--training time--number of parameters through three 2D projections, (b) discovery of optimal architectures using different preference functions, and (c) the optimal architectures and their performance, selected by the preference functions $p^1$, $p^2$, $p^3$ and $p^4$.}\label{Figure_example1}
\end{figure}
Accurate glucose trajectory forecasting using continuous glucose monitor (CGM) data is necessary for glycemic control. Using the GlucoBench dataset~\cite{sergazinov2024glucobench}, we design optimal architectures guided by preference functions. The GlucoBench dataset included the CGM measurements from five patients, and was pre-processed in Ref.~\cite{sergazinov2024glucobench}. In this example, we analyze the glucose trajectory of the first patient. The total length of the signal is 2,125 time steps, in which the initial 1,650 time steps are used for training and validation. Forecasting is performed using a reference time point at the 1,650th time steps. Using a sliding window technique, the datasets include 1,377 training samples and 154 validation samples. For each sample, the lookback window is 96 time steps and the forecasting window is 24 time steps. With the time interval of 5 minutes, the forecasting horizon is 2 hour, which is a good target time horizon for diabetes patients \cite{deng2022patient}. The models are trained on NVIDIA GeForce RTX 3090 and the implementation is in PyTorch~\cite{paszke2019pytorch}.

In this application, we consider both the number and the sequence of each type of blocks in the construction of architectures. 
Moreover, we focus on some particular combinations in the construction, though the possible combinations are theoretically infinite. In particular, for the composite model, the vector $\textbf{x}_1=[n, m, j, k]$ represents the number of each type of blocks, where $n, m, j$, and $k$ can each take values form 0 to 2. To ensure valid configurations, the case $\textbf{x}_1=[0, 0, 0, 0]$ is removed. The sequence of blocks is determined by $x_2$, which is chosen from the set $\{1, 2, 3, 4, 5, 6\}$. Each value of $x_2$ corresponds to a specific block arrangement:
\begin{itemize}
    \item $x_2=1$: {[}SSM, Attention, GRU, LSTM{]}
    \item $x_2=2$: {[}Attention, SSM, GRU, LSTM{]}
    \item $x_2=3$: {[}SSM, GRU, Attention, LSTM{]}
    \item $x_2=4$: {[}GRU, Attention, SSM, LSTM{]}
    \item $x_2=5$: {[}Attention, GRU, SSM, LSTM{]}
    \item $x_2=6$: {[}GRU, SSM, Attention, LSTM{]}
\end{itemize}
Finally, the hidden dimension of each block, denoted as $x_3$, is selected from $\{16, 32, 64\}$. This leads to a nominal total of $x_1\times x_2\times x_3=80\times6\times3=1440$ architectures. However, many of these configurations are duplicates because the order of blocks with identical presence/absence does not alter the final architecture. For example, [GRU=1, Attention=0, SSM=1, LSTM=1] is equivalent to [GRU=1, SSM=1, Attention=0, LSTM=1]. After systematically removing such duplicates, the final search space consists of 708 unique neural architectures. 

The architectures are evaluated via a three-dimensional objective functions $f = (f_1,f_2,f_3)$. Domain experts may select the specific metrics that are most relevant for their applications. In this study, we set $f_1$ to be the relative $\mathcal{L}_2$ error, $f_2$ to be the total training time, and $f_3$ to be the number of parameters. These metrics reflect both predictive performance and practical deployability. In glucose monitoring scenarios, models are often intended for use on wearable or edge devices with limited memory, therefore parameter count serves as an important indicator. Training time is treated as a separate optimization objective because it directly affects the efficiency of the architecture search and is not fully determined by model size. In parallel search settings, the wall clock time of each iteration is governed by the slowest model, so per-model training time influences the overall search cost. Moreover, architectures with similar parameter counts may exhibit substantially different training speeds. For this reason, we jointly optimize training time and parameter count. To examine this alternative perspective, Supplementary Information Section 8 reports a multi-objective optimization using relative $\mathcal{L}_2$ error, training time, and inference time.
Among the 708 candidate architectures, there are 8 Pareto optimal ones. 
\autoref{Figure_example1}(a) visualizes the Pareto front using three two-dimensional projections, each illustrating the relationships among training time, relative $\mathcal{L}_2$ error, and the number of training parameters.
In particular, the blue points in \autoref{Figure_example1}(a) represent the performance of all the neural architectures we constructed, while the red points correspond to the performance of Pareto optimal ones. 
This figure clearly shows that different performance criteria lead to different optimal architectures, and that no single architecture is optimal across all criteria. 

We propose a preference function-based approach to discover the optimal architecture.  
In particular, we start with the weighted sum preference functions. Note that, without loss of generality, we consider preference functions that may apply to the original data $f$ or re-scaled data $\hat{f}$,  where $\hat{f}$ is obtained by applying an increasing linear map $r$ on $f$.  
\begin{pref}
 $p^1 ({f}) := 1/3\hat{f}_1 + 1/3\hat{f}_2 + 1/3\hat{f}_3$. This preference function represents an equally weighted sum of the three objectives.
\end{pref}
\begin{pref}
$p^2({f}) := \hat{f}_1$.  This preference function is intended to find the optimal architectures, which obtains the smallest relative $\mathcal{L}_2$ error. 
\end{pref}
\begin{pref}
$p^3({f}):=\hat{f}_2$.  This preference function is intended to find optimal architectures that require the least training time. 
\end{pref}
\begin{pref}
$p^4({f}): = 3/5\hat{f}_1 + 1/5\hat{f}_2 + 1/5\hat{f}_3$. This preference function greatly emphasizes the accuracy while still considering training time and the number of parameters.
\end{pref}

We present optimal architectures discovered by these preference functions in~\autoref{Figure_example1}(b). 
In particular, we highlight with star symbols the selected optimal points, which correspond to the performance of architecture that best matches the specified preference function. 
Note that the coordinates in ~\autoref{Figure_example1}(b) are the normalized objective functions. 

The preference function $p^1$ selects an architecture with one GRU block, which achieves a moderate relative $\mathcal{L}_2$ error of $6.79\%$, a low training time of 8.212 s, while the number of parameters is 2,201. The preference function $p^2$ yields an architecture with the best accuracy with the relative $\mathcal{L}_2$ error of $2.37\%$, but it also requires the highest cost. This designed architecture includes a mix of one multi-head Attention block and two GRU blocks. We also independently fine-tune the Transformer, which achieves the relative $\mathcal{L}_2$ error of 4.10$\%$.  The architecture corresponding to the preference function $p^3$ addresses efficiency, which chooses a configuration with a single-layer GRU, leading to the shortest training time. Finally, $p^4$ prioritizes accuracy while also taking training costs into consideration. The selected architecture includes one multi-head attention block and two GRU blocks. The optimal architectures and their performance are summarized in \autoref{Figure_example1}(c). 

Another interesting question is that, given a particular Pareto optimal architecture, under what situation it is the optimal choice. 
This is associated with the rediscovery of preference functions for specific network architectures. 
For particular weighted-sum preference functions, 
we employ the Linear Programming (LP) method. 
Indeed, given a particular architecture $x$, one can formulate the problem of rediscovering as a constrained optimization (minimization) problem. The new objective function is the weighted sum of the performance of 
$x$, and the optimization is performed on the weights, subject to the constraint that this weighted sum should be less than or equal to the weighted sum of the performance of all other architectures.
 
We consider four representative network architectures selected from the Pareto optimal ones, and find their optimal weights:
\begin{itemize}
    \item \textbf{Case 1}: The architecture consists of one Attention block and one LSTM block, with a hidden dimension of 16. The weighted sum preference function coefficients are $\lambda_1 = 0.383$, $\lambda_2 = 0$ and $\lambda_3 = 0.617$.
    \item \textbf{Case 2}: The architecture includes one SSM block, one Attention block, and two GRU blocks, with a hidden dimension of 32. The coefficients are $\lambda_1 = 0.697$, $\lambda_2 = 0$ and $\lambda_3 = 0.303$.
    \item \textbf{Case 3}: The architecture contains only one GRU block and has a hidden dimension of 64. The corresponding coefficients are $\lambda_1 = 0.048$, $\lambda_2 = 0.952$ and $\lambda_3 = 0$.
    \item \textbf{Case 4}: This architecture is composed of two GRU blocks, one SSM block, one Attention block, and one LSTM block, with a hidden dimension of 32. The coefficients are $\lambda_1 = 1$, $\lambda_2 = 0$ and $\lambda_3 = 0$.  
\end{itemize} 
The details are summarized in Supplementary Information Section 1.1.
On the other hand, the coefficients also illustrate each architecture's specific advantages. In particular, Cases 2 and 4 represent composite architectures composed of various types of blocks. These cases identify significantly higher values for $\lambda_1$ compared to the other two coefficients, indicating that these architectures prioritize accuracy while maintaining lower training costs. Case 3 involves a single-layer GRU with a hidden dimension of 64, identifies a larger $\lambda_2$ and a smaller $\lambda_1$. These values imply that this architecture addresses a shorter training time and reduced accuracy but does take into account the number of parameters. Supplementary Information Section 9 demonstrates the comparison between our best architecture (lowest relative $\mathcal{L}_2$ error) and several established baselines (ARIMA, ETS-Holt-Winters, PatchTST~\cite{nie2022time}, Temporal Fusion Transformers~\cite{lim2021temporal}, TiDE~\cite{das2023long}, Chronos~\cite{ansari2024chronos}, and DLinear~\cite{zeng2023transformers}). These baselines are implemented using AutoGluon~\cite{shchur2023autogluon}, an open-source AutoML library for time series forecasting.

\subsection*{Forecasting significant wave heights in Pelabuhan Ratu, Indonesia}

\noindent
ERA5 hourly data~\cite{hersbach2023era5} are used to forecast significant wave height in this study. The forecast is carried out with a reference time of January 25, 2020, at 01:00:00. The observations of the significant wave heights from January 1, 2014, at 00:00:00 to January 25, 2020, at 01:00:00 are used as the training data. For each sample, the lookback window is 648 time steps (27 days) and the forecasting time is 72 time steps (3 days). The models for this example are trained on NVIDIA H100 GPU.

\begin{figure}[ht]
\centering
\includegraphics[width=\textwidth, trim = 0cm 11cm 0cm 0cm, clip]{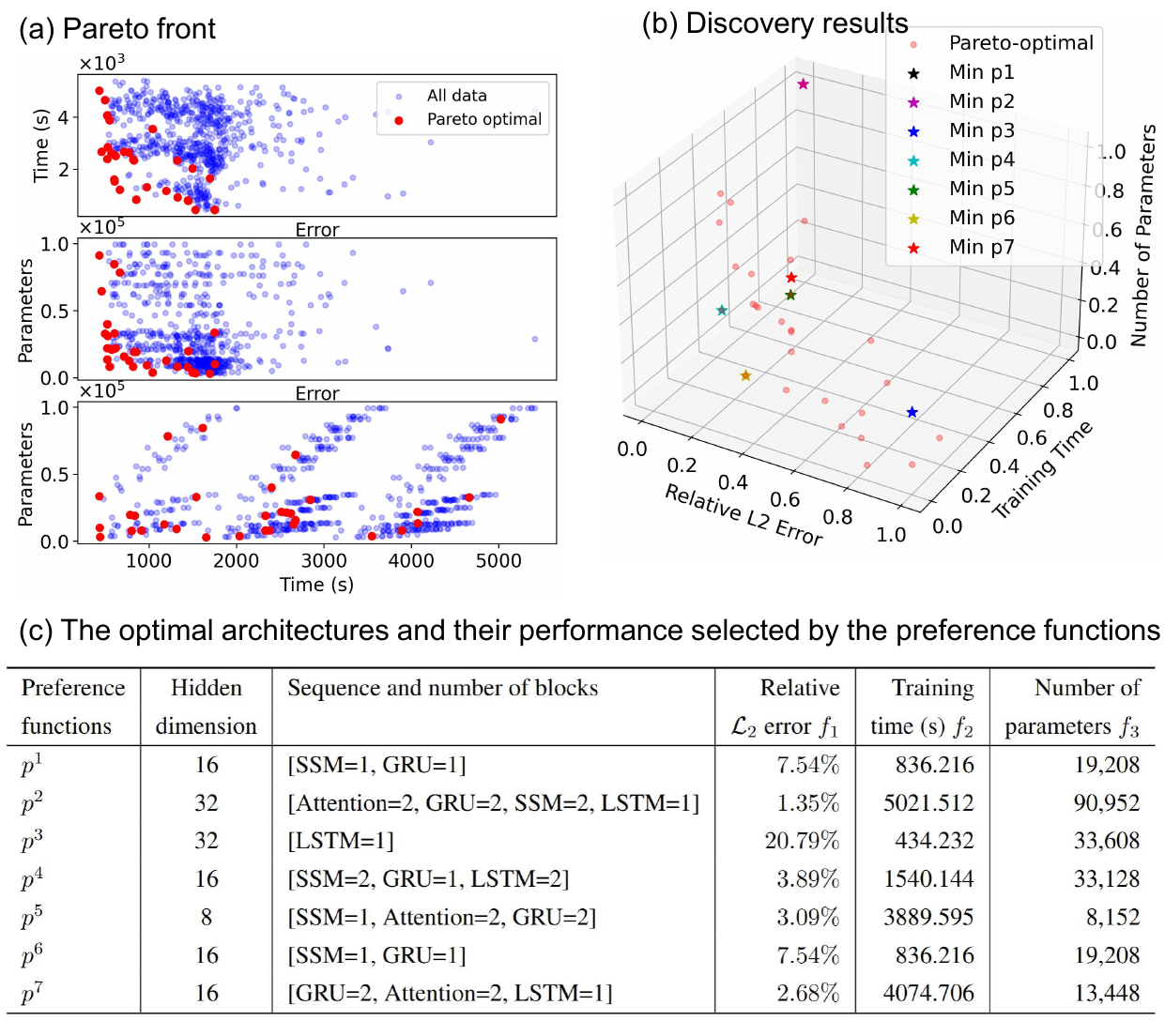}\\
\caption{Wave height prediction: (a) Pareto front showing relative error--training time--number of parameters through three 2D projections, (b) discovery of optimal architectures using different preference functions, and (c) the optimal architectures and their performance, selected by the preference functions $p^1$ -- $p^{7}$.}\label{Figure_example2}
\end{figure}

The construction of architectures in this application is similar with the first application, which considers both the number and the sequence of each type of blocks. However, the hidden dimension of each block $x_3$ is selected from $\{8, 16, 32\}$. This setup yields a total of 708 network architectures to forecast the significant wave height. The Pareto front is presented in \autoref{Figure_example2}(a), and there are 31 Pareto optimal architectures.

To discover optimal architectures, we consider the four weighted sum preference functions $p^1,p^2,p^3,p^4$, and three new nonlinear preference functions. 
\begin{pref}\label{pref5}
$p^5({f}) := p^5_{1}({f}_1) + \frac{1}{2}\hat{f}_2+ \frac{1}{2}\hat{f}_3$,  
where 
\begin{equation}\label{surge_p51}
    p^5_{1}(f_1) =
    \begin{cases} 
        0, & {f}_1  \leq 0.06 \ , \\ 
        10^3, &   {f}_1 > 0.06  \ .
    \end{cases}
\end{equation}
This preference function emphasizes the importance of the model accuracy by maintaining a relative $\mathcal{L}_2$ error below 0.06 to avoid penalty. However, once accuracy is guaranteed, minimizing the computational cost ($f_2$ and $f_3$) becomes the critical factor. 
\end{pref}

\begin{pref}\label{pref6}
$p^6({f}) := \hat{f}_1 + p^6_{2}({f}_2) + p^6_{3}({f}_3)$, where
\begin{equation}\label{surge_p62}
    p^6_{2}(f_2) =
    \begin{cases} 
        0, &  f_2 \leq 1,000 \ , \\ 
        10^3, & f_2 > 1,000 \ ,
    \end{cases}
\end{equation}
and
\begin{equation}\label{surge_p63}
    p^6_{3}(f_3) =
    \begin{cases} 
        0, & f_3 \leq 35,000 \ , \\ 
        10^3, &  f_3 > 35,000 \ .
    \end{cases}
\end{equation}
This preference function imposes hard constraints on the computational costs, which limits the maximum training time to be 1,000 s, and the maximum number of parameters to be 35,000. 
Among the fast models, with values lower than these thresholds, the model accuracy is prioritized. 
\end{pref}
\begin{pref}\label{7}
$p^7(f):= 0.7 \hat{f}_1 + 0.01  \log_2(f_2)  + 0.01  \max(0, f_3 - 20,000) $. 
In this preference function, the model accuracy $f_1$ is heavily weighted to emphasize model performance. The penalty on the training time is reduced, and is represented by the logarithm of the training time. The number of parameters is penalized linearly when it exceeds 20,000.
\end{pref}

We present the discovery of optimal architectures by the preference functions $\{p^i \}_{i \in \{1,\dots,7 \}}$ in \autoref{Figure_example2}(b).
The configurations of the designed network architecture and performance metrics are summarized in~\autoref{Figure_example2}(c). The nonlinear preference function $p^5$ selects architectures with one SSM layer, two Attention layers and two GRU layers, which achieves the relative $\mathcal{L}_2$ error below $6\%$. The $p^6$ chooses the same architecture with $p^1$. Comparing \autoref{Figure_example1}(c) and \autoref{Figure_example2}(c), the designed architectures in \autoref{Figure_example2}(c) are more complex than those in \autoref{Figure_example1}(c), likely due to the increased complexity of the current problem. The rediscovery results for this example are summarized in Supplementary Information Section 1.2.

\subsection*{Predicting Vortex-Induced Vibrations (VIV) of marine risers}
\noindent
The Norwegian Deepwater Programme (NDP) conducted VIV tests on a $38$ m long riser. In this study, we focus on the cross-flow strain recorded by the 10$^{th}$ strain gauge under a uniform flow with a speed 0.7 m$/$s. The total length of the signal is 156,600 time steps. The stationary part with the 30,000 time steps is chosen in this example, in which the first 27,000 time steps are used for training and validation. The goal of this example is to forecast 60 time steps from the current time defined at 27,000 s. A sliding window technique is applied by using a window length of 1,000 time steps with a shift interval of 5 time steps, which results in 5,201 samples in total. Here, 90$\%$ of these samples are chosen as the training samples while the rest  are used for validation. For each sample, the lookback window is 900 time steps. The models for this example are trained on NVIDIA RTX A6000.

\begin{figure}[ht]
\centering
\includegraphics[width=\textwidth, trim = 0cm 11cm 0cm 0cm, clip]{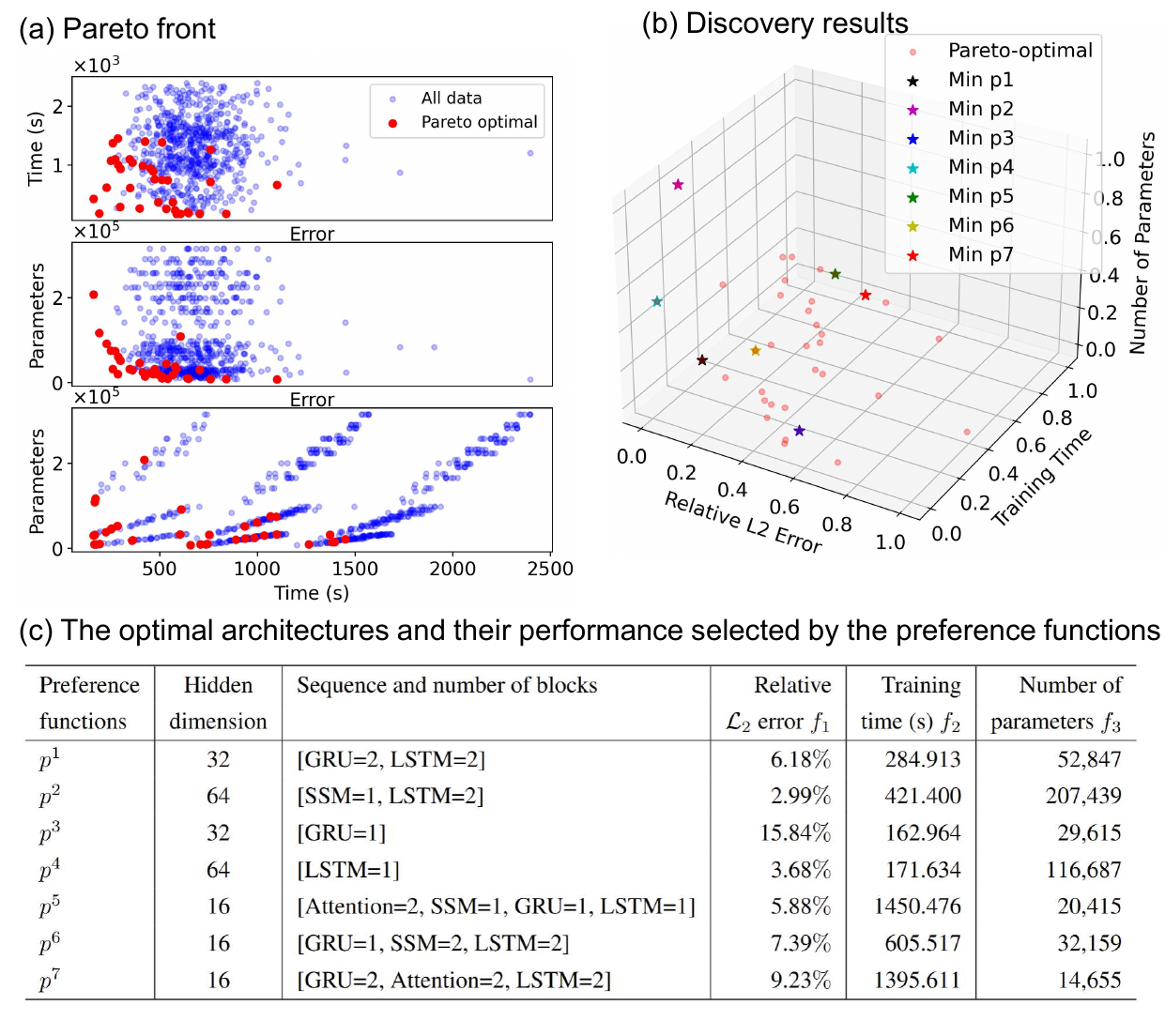}\\
\caption{Marine risers: (a) Pareto front showing relative error--training time--number of parameters through three 2D projections, (b) discovery of optimal architectures using different preference functions, and (c) the optimal architectures and their performance, selected by the preference functions $p^1$ -- $p^{7}$, for VIV predictions.}\label{Figure_example3}
\end{figure}

\begin{figure}[htbp!]
\centering
\includegraphics[trim=0cm 5cm 0cm 0cm,clip=true,width=0.7\textwidth]{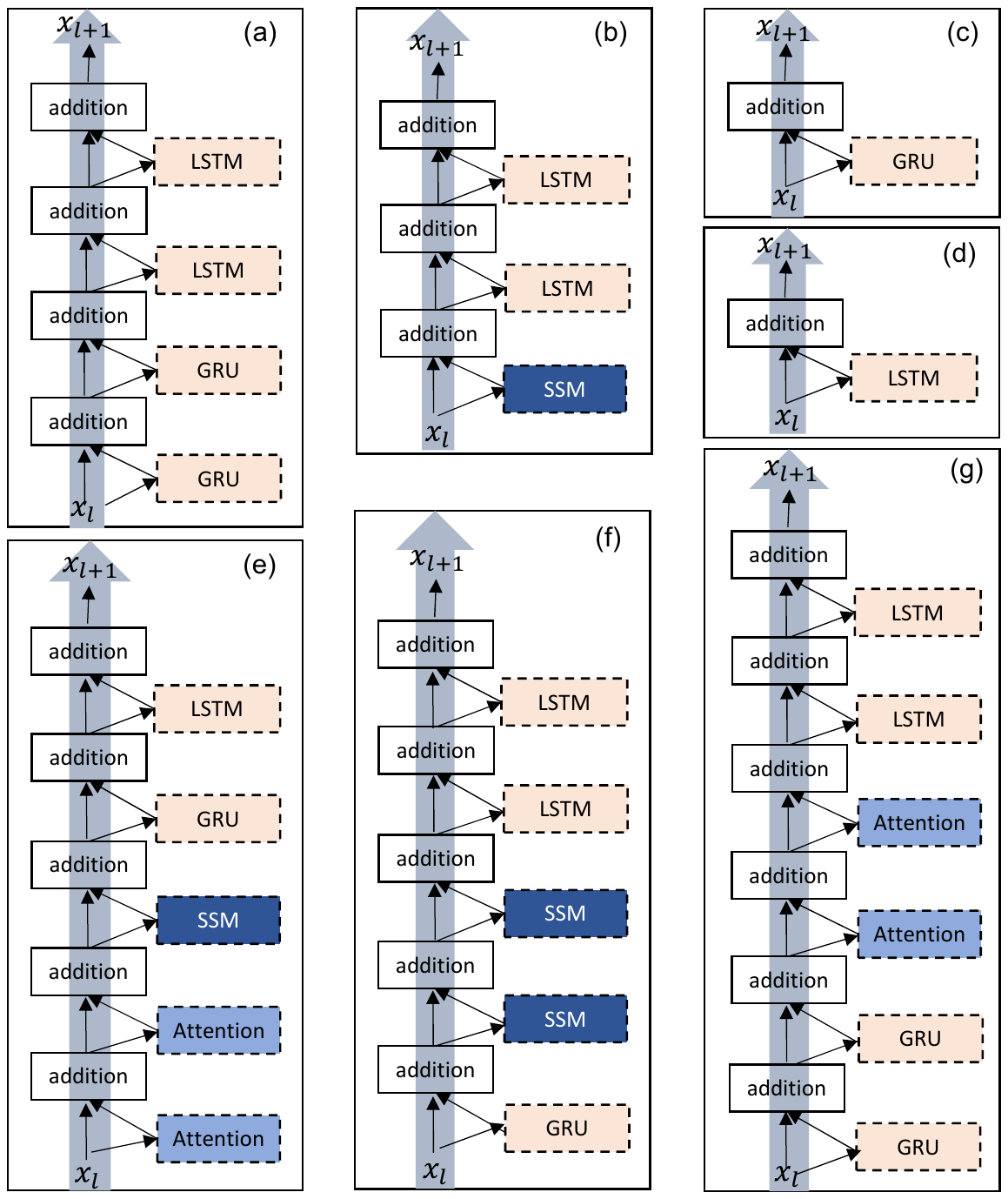}\\
\caption{Designed network architectures for marine risers based on given preference functions: (a) $p^1$, (b)  $p^2$, (c) $p^3$, (d) $p^4$,  (e) $p^5$, (f)  $p^6$, (g) $p^7$.}\label{designed_NN_riser}
\end{figure}

The construction of architectures in this application is the same with the first application, which considers both the number and the sequence of each type of blocks. This setup yields a total of 708 network architectures to forecast the motion of the riser. The Pareto front is presented in \autoref{Figure_example3}(a), and there are 35 Pareto optimal architectures. 
To discover optimal architectures, we consider the four weighted-sum preference function $p^1,p^2,p^3,p^4$, and the three nonlinear preference functions $p^5,p^6,p^7$.  
We present the optimal architectures discovered by these preference function in the Pareto front in \autoref{Figure_example3}(b). The network architectures and performance metrics are also summarized in~\autoref{Figure_example3}.  
We present detailed illustrations of the second modules of the designed architectures in \autoref{designed_NN_riser}. The nonlinear preference function $p^5$ has a threshold of relative $\mathcal{L}_2$ error equals $6\%$, which finds the architecture with two Attention blocks first, followed by one SSM block, one GRU block and one LSTM block (see \autoref{designed_NN_riser}(e)). Here, $p^6$ has thresholds for the training time and the number of parameters to ensure the computational efficiency. The resulting architecture includes a sequence starting with one GRU block, followed by two SSM blocks, and concluding with two LSTM blocks (see \autoref{designed_NN_riser}(f)). As shown in \autoref{designed_NN_riser}(g), the designed architecture for $p^7$ consists of a sequence starting with two GRU blocks, followed by two Attention blocks and two LSTM blocks to guarantee the computational accuracy and computational costs. Moreover, the rediscovery results are summarized in Supplementary Information Section 1.3.

\subsection*{Predicting the motions of floating offshore structures}
\noindent
In this example, we study a semi-submersible platform with the mooring system. The parameters of the platform and the simulation details are summarized in the Supplementary Information Section 2. The aim of this example is to forecast the surge motion based on the historical data. To train the proposed composite model, we conduct a simulation lasting about 3 hours on a singe CPU. The data collected from the beginning of the simulation 0 s to 10,373 s are used for training and validation. The forecasting time is 12 s from the current time (10,373 s). We further apply a sliding window technique by using a window length of 1,000 time steps per sample with a shift interval of 10 time steps. This process yields 5,100 samples and each comprises 1,000 time steps. The time interval is 0.2 s.
For model training, 90$\%$ of these samples are randomly chosen, while the remaining $10\%$ are allocated for validation. The models are trained on NVIDIA GeForce RTX 3090.

\begin{figure}[ht]
\centering
\includegraphics[width=\textwidth, trim = 0cm 11cm 0cm 0cm, clip]{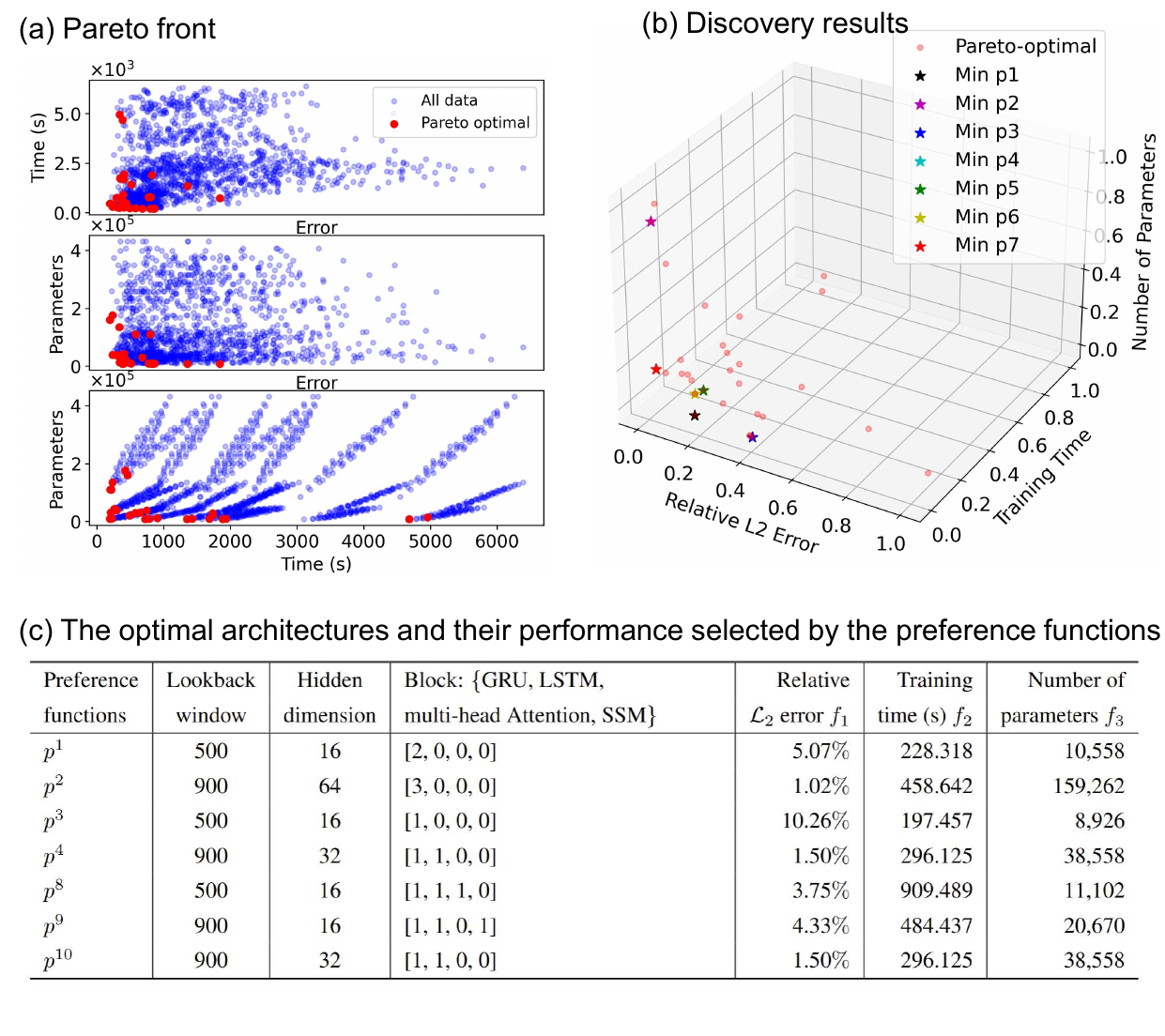}\\
\caption{Surge motion of a platform: (a) Pareto front showing relative error--training time--number of parameters through three 2D projections, (b) discovery of optimal architectures using different preference functions, and (c) the optimal architectures and their performance, selected by the preference functions $p^1$ -- $p^4$ and $p^8$ -- $p^{10}$.}\label{Figure_example4}
\end{figure}

In this example, only the number of each type of block is considered. The range of each element in the vector $\textbf{x}_1$ is extended from $[0,2]$ to $[0,3]$. To ensure valid configurations, the combination $\textbf{x}_1=[0, 0, 0, 0]$ is excluded, resulting in a total of $255$ possible network architectures. The block sequence is fixed to [SSM, Attention, GRU, LSTM]. The lookback window, denoted by $x_3$, is selected from $\{500, 900\}$, and the hidden dimension, $x_4$, is chosen from $\{16, 32, 64\}$. The down-sampling is fixed to 2. To reduce the number of parameters, all attention blocks share the same parameters, except in Example 1. This example illustrates the effect of using distinct parameters in different attention blocks. This setup yields a total of 1,530 cases, where there are 30 Pareto optimal ones. We present the Pareto front in~\autoref{Figure_example4}(a).  

For the surge motion, four weighted sum preference function $p^1,p^2,p^3,p^4$ and three nonlinear preference functions are considered in this example. The nonlinear preference functions $p^8, p^9, p^{10}$ have similar forms as $p^5, p^6, p^7$, but the threshold values of the evaluation metrics are changed. 
\begin{pref}
$p^8(f) := p^8_1(f_1) + \frac{1}{2}\hat{f}_2+ \frac{1}{2}\hat{f}_3$, where 
\begin{equation}\label{riser_p81}
    p^8_1(f_1) =
    \begin{cases} 
        0, &  f_1 \leq 0.04 \ , \\ 
        10^3, &  f_1 > 0.04 \ .
    \end{cases}
\end{equation}
This preference function has a similar form as $p^5(f)$. Here, we change the threshold values on the evaluation of the relative error. 
\end{pref}
\begin{pref}
$p^9(f) = \hat{f}_1 + p^9_{2}(f_2) + p^9_{3}(f_3) $, where 
\begin{equation}\label{riser_p92}
    p^9_{2}(f_2) =
    \begin{cases} 
        0, &  f_2 \leq 500 \ , \\ 
        10^3, &  f_2 > 500 \ .
    \end{cases}
\end{equation}
and
\begin{equation}\label{riser_p93}
    p^9_{3}(f_3) =
    \begin{cases} 
        0, &  f_3 \leq 35,000 \ , \\ 
        10^3, &  f_3 > 35,000 \ ,
    \end{cases}
\end{equation}
This preference function has a similar form as $p^6(f)$. Here, we change the threshold values on the evaluation of the training time and the number of parameters. 
\end{pref}
\begin{pref}
    $p^{10}(f) := 0.7 \hat{f}_1 + 0.001  \max(0, f_2 - 500) + 0.01  \log_2(f_3) $. In this preference function, the model accuracy $f_1$ is heavily weighted to emphasize model performance. The training time $f_2$ is also linearly penalized when the training time exceeds 500 s. To reduce the penalty's impact on larger models, the model complexity $f_3$ is represented by the logarithm of the number of parameters.
\end{pref}

\autoref{Figure_example4}(b) plots the discovery of optimal architectures by the preference functions. 
The configurations of the designed network architecture and performance metrics are summarized in~\autoref{Figure_example4}(c).
We apply the Linear Programming method to rediscover the optimal architectures, for weighted sum preference functions. The results are summarized in Supplementary Information Section 1.4. Moreover, comparing our best-performing model and the LSTM fine-tuned using Optuna~\cite{optuna_2019} is an interesting direction to show the advantages of our method, which are summarized in Supplementary Table S7 and Table S8.

\subsection*{Iterative architecture samplings for glucose prediction task}
\begin{figure}[ht]
\centering
\includegraphics[width=\textwidth, trim = 0cm 3cm 0cm 0cm, clip]{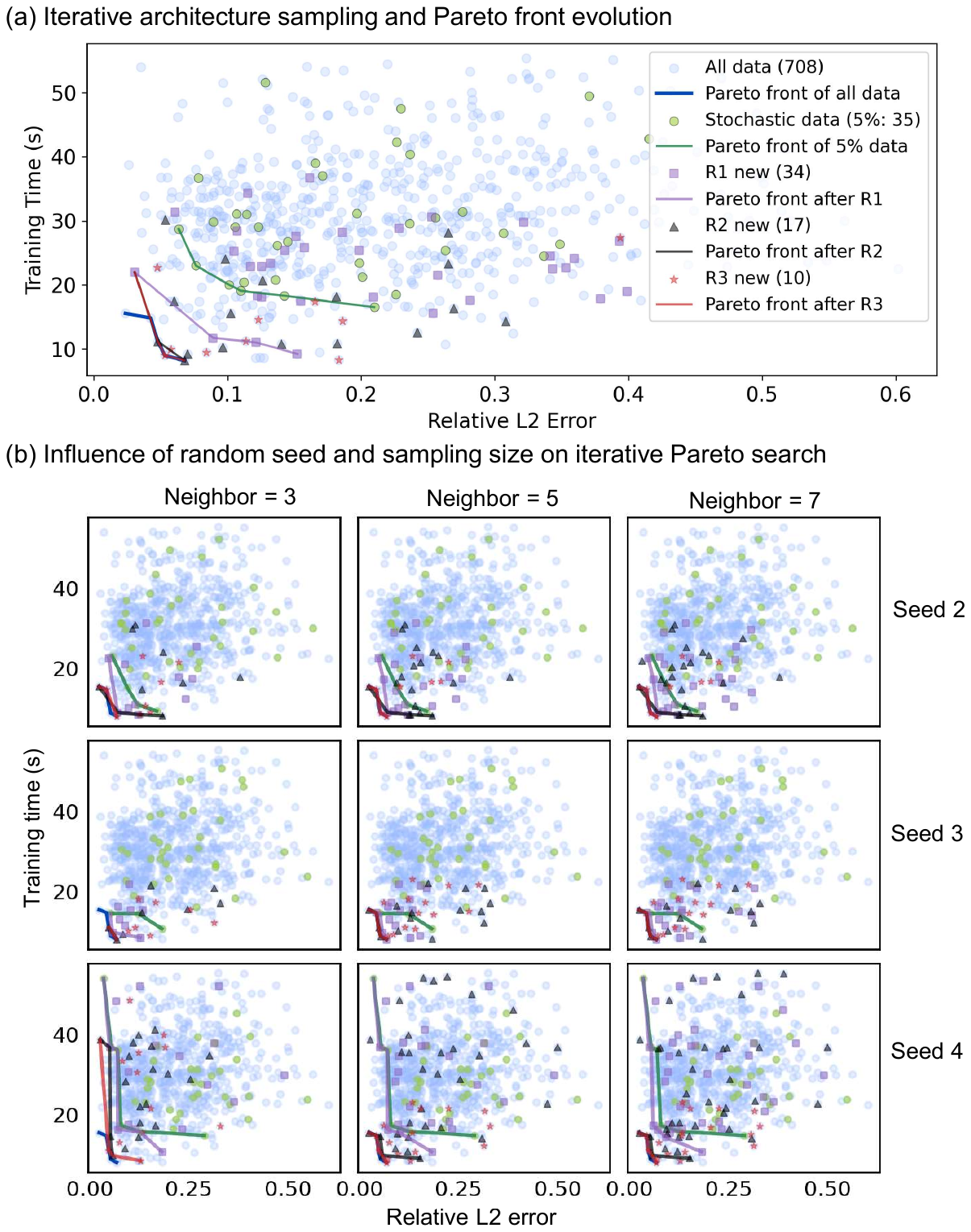}\\
\caption{A sampling approach to reduce the training time of optimal architecture search: (a) Iterative architecture sampling and Pareto front evolution, (b) Robustness analysis of the iterative architecture sampling.}\label{Figure_sampling}
\end{figure}

\begin{figure}[ht]
\centering
\includegraphics[width=\textwidth, trim = 0cm 0cm 0cm 0cm, clip]{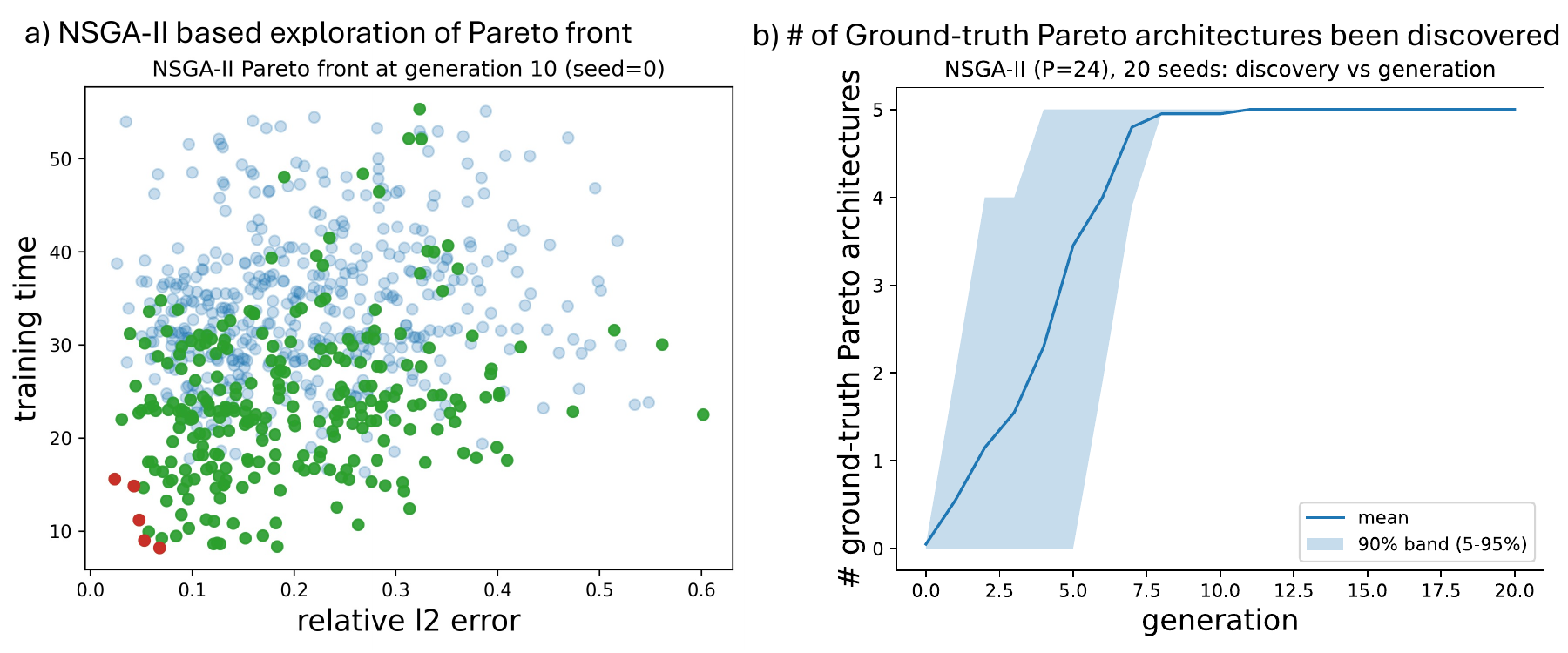}\\
\caption{ 
Budgeted NSGA-II baseline for exploring the Pareto front in the training-time vs.\ relative $\mathcal{L}_2$-error space for the glucose prediction task.
(a) Representative run (seed $=0$): light blue dots show all $708$ architectures; green dots are the architectures evaluated by NSGA-II up to generation $10$; red dots are the Pareto optimal subset within the evaluated set (NSGA-II’s current Pareto approximation).
(b) Discovery of ground-truth Pareto-optimal architectures over generations, aggregated over $20$ random seeds and reported as mean with a $90\%$ band (5--95\% quantiles).
}\label{Figure_nsga}
\end{figure}

To evaluate the effectiveness of the iterative architecture sampling methods, we use the glucose prediction task as an example. \autoref{Figure_sampling} and \autoref{Figure_nsga} demonstrate the resulting Pareto fronts by the neighborhood sampling approach and NSGA-II procedure (see, e.g.,~\cite{deb2002fast,lu2019nsga,lu2020nsganetv2}), respectively.

\autoref{Figure_sampling}(a) shows how the Pareto front progressively evolves with each sampling round. The light blue points represent all previous trained architectures (708 in total), and the dark blue curve denotes their Pareto front, which serves as the ground truth in the performance space. In this new example, Pareto optimal points are identified based on two performance metrics, the training time and the relative $\mathcal{L}_2$ error. We begin by randomly sampling 5\% of the architectures (35 cases, shown as green points) to start the search. The corresponding Pareto front (green line) forms the initial approximation. Each point on this front corresponds to a particular optimal architecture in the architectural design space. To refine the front, we iteratively expand the search by sampling new candidate architectures around the current Pareto-optimal ones. Specifically, for each Pareto-optimal architecture $X$, we randomly select several neighboring architectures within the architectural space, while enforcing constraints to avoid duplicates. In the first iteration, 34 new architectures (purple squares) are added, and the updated Pareto front (purple line) shifts closer to the ground truth. The second iteration follows the same procedure, producing 17 new architectures (black triangles) and a further refined front (black line). The third iteration adds red-star points, which yields a red Pareto front that nearly coincides with the ground truth. This progressive evolution clearly demonstrates that as the algorithm iteratively explores local neighborhoods of the Pareto-optimal architectures, the front rapidly converges toward the true Pareto front. After only three iterations (96 architectures in total), the Pareto front approximates that of all 708 architectures, reducing the total training cost by nearly eightfold.
\autoref{Figure_sampling}(b) evaluates the robustness of the method under different random seeds and neighborhood sizes. The random seed determines the initial 5\% subset, while the number of neighbors controls the extent of local exploration. Experiments with neighbor counts of 3, 5, and 7 under three additional random seeds show consistent convergence behavior. Using 5 neighbors achieves a Pareto front comparable to that obtained with 7 neighbors but with fewer total samples. Across all cases, the iterative sampling strategy maintains the same evolutionary pattern, where the Pareto front gradually converges to the ground truth while substantially reducing the overall search cost.

\autoref{Figure_nsga} shows how NSGA-II recovers the true Pareto-optimal architectures as generations progress. NSGA-II is initialized with a population of size $P=24$ (about $3.4\%$ of the full space). At each generation, it performs non-dominated sorting and crowding-distance based selection, and generates up to $P$ duplicate-free offspring via discrete crossover and mutation over the architectural design variables (with a canonicalization step to avoid redundant evaluations). We run NSGA-II for a fixed number of generations and track the set of all architectures evaluated up to each generation. \autoref{Figure_nsga}(a) visualizes a representative run.
Light blue dots indicate the performance of all $708$ architectures, while green dots denote the subset evaluated by NSGA-II up to the displayed generation.
Red dots correspond to the Pareto-optimal subset within the evaluated set, which constitutes the current approximation of the Pareto front.
In this representative run, NSGA-II recovers all $5$ ground-truth Pareto-optimal architectures after evaluating $116$ distinct architectures (generation $4$ under an early-stopping criterion), and by generation $10$ the recovered front is stable and matches the ground-truth Pareto set.
To quantify robustness, we repeat the procedure for $20$ random seeds and measure the sample complexity, defined as the number of distinct architectures that must be evaluated (i.e., fully trained) until all $5$ ground-truth Pareto-optimal architectures are discovered.
Across $20$ runs, NSGA-II requires $166.15$ evaluations on average (median $172.5$), with a $90\%$ interval of $[116,207]$ and a best/worst case of $107$/$274$ evaluations.
\Cref{Figure_nsga}(b) reports the mean discovery trajectory over generations together with a $90\%$ uncertainty band.

We have CGM measurements from five patients, and the optimal architecture identified in the glucose example is based on Patient 1’s trajectory. Because all five trajectories arise from the same domain, we hypothesize that the optimal architecture found for one trajectory may generalize to others. This motivates an architecture-transfer strategy that avoids running a full search for each patient. We begin by directly transferring Patient 1's optimal architecture to a target trajectory, which is used as the starting point rather than reinitializing a global search. To adapt this architecture to patient-specific dynamics, we introduce a lightweight refinement procedure that explores only a small neighborhood of the architecture space. The refinement proceeds in rounds, where each round samples several neighboring architectures obtained by localized modifications of the current best model. These candidates are evaluated on the target trajectory, and the one with the lowest error becomes the best model for the next round. Across three rounds, the procedure evaluates less than 15 architectures in total, compared with 708 models in the original search space, yielding a reduction in computational cost approaching two orders of magnitude.
Table~\ref{TF_diff_patient} summarizes the results. After three refinement rounds, errors decrease by approximately 34$\%$–74$\%$ relative to direct transfer, and the refined architectures closely match the ground-truth optima obtained from full search. This demonstrates that our transfer-and-refine strategy provides accurate predictions with dramatically reduced search effort. The full procedure and implementation details are provided in the Supplementary Information Section 7.
\begin{table}[h!]
\centering
\caption{Evaluation of architecture transfer across same-domain trajectories. ``Ground-truth error" is the error of the globally optimal architecture obtained from a full search on each patient's data. The starting architecture is the optimal model selected for Patient 1 ([Attention=1, GRU=2]) with width 32). ``Start error" is the error from directly transferring Patient 1’s architecture to other patients. ``Best architecture after 3 Rs" and ``Error after 3 Rs" denote the lowest error architecture identified after three rounds of random sampling. ``Improvement" measures the reduction from the start error to the refined error after three rounds.
}\label{TF_diff_patient}
\resizebox{\textwidth}{!}{%
\begin{tabular}{l  l l l l l l}
\hline
\specialrule{0.3pt}{0pt}{0pt}
Patient &Ground-truth & Start & Best architecture  & Width & Error & Improvement \\
 &      error & error & after 3 Rs & & after 3 Rs & \\
\hline
P2 & 0.0207&  0.1165 & [Attention=2, GRU=2] & 16 & 0.07472 & +35.87\%  \\
P3 & 0.0364&  0.05836 & [SSM=2, Attention=1, GRU=1] & 32& 0.0385  & +34.03\%  \\
P4 & 0.0175 &  0.08634 & [GRU=1, LSTM=2] & 32 & 0.02475 & +71.34\% \\
P5 & 0.0110&  0.07558 & [LSTM=1]  & 32 & 0.01952 & +74.17\%  \\
\hline
\specialrule{0.3pt}{0pt}{0pt}
\end{tabular}}
\end{table}

\bigbreak

\noindent

\section*{Discussion}
We proposed a general framework to automatically discover composite neural architectures, which integrated GRU, LSTM, multi-head Attention, and SSM blocks. Compared with traditional models, which have a fixed architecture, the proposed model is more flexible. The number and sequence of each block are designed by using the multi-objective optimization approach to satisfy specific engineering needs. As an inverse direction of designing the best network architecture, the particular preference function for a given Pareto optimal network architecture is also identified by the Linear Programming method. 

For benchmarks, we considered four diverse real-world applications. We observed several interesting phenomena through cross-data comparison: (1) When the objective was solely to minimize training time, the simplest architectures, such as a single-layer GRU or LSTM, consistently performed the best. (2) When the objective shifted to minimizing the relative $\mathcal{L}_2$ error or optimizing combined evaluation metrics, the optimal architecture became highly dependent on the dataset. (3) When a balanced compromise was preferred (i.e., $p^1$), the optimal architectures were of moderate complexity. These typically include two different types of blocks, which suggests that neither overly simple nor overly complex designs provide the best overall balance. (4) If the highest prediction accuracy was prioritized, the optimal architectures were generally more complex and composed of multiple block types, which indicates that accuracy-driven designs benefit from architectural diversity and depth. (5) Across more experiments, GRU blocks were the most frequently used component. For the preference function $p^1$, at least one GRU block was always included. Additionally, for the preference function $p^2$, the optimal architectures in most datasets also included GRU blocks, demonstrating their consistent contribution to both balanced and accuracy-focused performance. (6) Across all orders, the relative $\mathcal{L}_2$ error distributions share similar shapes, which indicates that model performance is primarily governed by the composition of block types rather than their sequence. 
Note that our contributions do not only come from introducing a composite model but also  from proposing an important multi-objective optimization method to evaluate and design network architectures based on the preference function. Taken together, our results explain why the literature presents different conclusions about the performance of different models that are based on fixed neural architectures. 

In our experiments, we observed that the offline computational cost is dominated by training non–Pareto-optimal architectures, which account for the vast majority (around $95\%$) of all architectures constructed during the search. This makes the brute-force exploration of the design space prohibitively expensive. 
These observations motivate the question: can we first gather coarse information about the Pareto front, and then restrict the search to composite architectures that lie ``near'' the currently Pareto-optimal ones, thereby reducing the offline cost? 
We introduce two iterative sampling procedures for direct exploration of the architecture space. The first, a neighborhood sampling approach, refines the Pareto front by sampling new candidates near the current optimal set at each iteration. The second utilizes NSGA-II to ensure exploration and diversity through evolutionary operators. While we focus on these two methods, our framework is modular and compatible with other established multi-objective search procedures, which can be used as a refinement stage on top of candidate sets produced by global NAS pipelines.  
Numerically, our first iterative sampling achieves a strong efficiency gain: with three refinement rounds, evaluating only about $13.5\%$ of all candidates, it already yields a high-quality approximation of the full Pareto front, consistently across random seeds. 
Our approach is specifically tailored for the dense refinement of architectures within constrained local search spaces rather than global search over exponentially large domains. We position this method as a potentially powerful post-refinement stage that complements existing pipelines by taking a set of high-performing candidates and extracting a reliable Pareto-optimal subset based on objectives.

An interesting further direction is to predict the performance function with respect to parameters. However, preliminary numerical experiments indicate that the performance matrix exhibits highly irregular behavior. This irregularity underscores the need for optimization tools that can select high-performing architectures on demand across different scenarios. Uncertainty quantification is also an important future direction. When uncertainties are taken into account, the performance metrics are no longer fixed values but instead within bounded regions, which cause the Pareto front to appear as a volume region in the performance space. Consequently, the notion of a single optimal architecture is replaced by an optimal architecture region. Another promising avenue is to integrate recent agentic optimization techniques~\cite{georgiev2025mathematical,jiang2025agenticsciml}, such as AlphaEvolve-style evolutionary agents, into our framework. In such a setting, an autonomous agent would iteratively propose new architectures, query their performance, update an internal surrogate of the performance landscape, and adapt its sampling strategy to focus on promising Pareto regions. Combining these agentic strategies with our Pareto-aware sampling procedure could further reduce offline computational cost and enable more efficient adaptive exploration of large composite-architecture spaces.

\section*{Methods}\label{sec-methods}
\subsection*{Composite architectures}
We propose to search for composite architectures, which are hybrids consisting of GRU, LSTM, multi-head Attention, and SSM blocks to leverage the best features of each while pursuing the objectives that a specific engineering or another application needs. The number and the sequence of each type of block are tunable parameters, which will be obtained based on multi-objective optimization method, aiming to design the best architectures tailored to specific engineering needs and target objectives.

The model architecture is shown in \autoref{Figure_architecture}(a), which includes three main modules. Two linear transformations can be used in the first (Embedding) and the third (LN) modules, which lift the input to a higher dimensional representation and project the output to the original dimension, respectively. The second module is the main step, which hybridizes GRU, LSTM, multi-head Attention, and SSM blocks. Note that \autoref{Figure_architecture}(a) only demonstrates one possible scenario about the number  and sequence of each block. A plurality of architectures can be created based on this composite model. 
Each LSTM block includes a hidden state and a cell state, which are updated during each time step. The GRU block is similar to the LSTM block but includes fewer gates so that fewer parameters are used. The readers are referred to Refs.~\cite{hochreiter1997long,cho2020learning} for the details of these architectures. \autoref{Figure_architecture}(a) also demonstrates the architectures of a Transformer block and a SSM block, which consist of two sequential residual modules. In the first module, the input undergoes layer normalization first. Then, we apply a multi-head Attention or SSM mechanism to enable the model to focus on the input sequence in parallel. The resulting output is then added to the original input by a residual connection. Similarly, the output of the first module passes a layer norm and a feed-forward neural network (FFN) in the second module, whose output is added to the module's input by another residual connection. Note that both the Transformer and SSM blocks use the Pre-LN model, which puts the layer normalization inside the residual blocks and converges much faster than the traditional method~\cite{xiong2020layer}.

To design the optimal architecture based on the composite model and the multi-objective optimization method, the parameterized spaces of different architectures are considered. The number of each block is defined as 
a vector $\textbf{x}_1=[n, m, j, k]$, where $n$, $m$, $j$ and $k$ represent the number of GRU, LSTM, multi-head Attention and SSM blocks, respectively. The sequence of blocks is represented by $x_2$, which is chosen from a discrete set of integers $\{1, 2, \cdots, p-1, p\}$. Each integer represents a specific sequence. For example, the sequence of blocks is [SSM, Attention, GRU, LSTM] when $x_2=1$ and it is [Attention, SSM, GRU, LSTM] when $x_2=2$. The integer $p$ represents the total number of sequences, which will be determined based on the specific problem we solve. By combining GRU, LSTM, multi-head Attention, and SSM blocks, and considering the number and sequence of blocks as tunable parameters, the proposed model not only includes many existing architectures but also can discover a wide range of new and optimal architectures.

\subsection*{Multi-Objective optimization and Pareto optimal architecture discovery} 
In this section, we give details of the multi-objective optimization methodology for selecting the optimal architectures.  In particular, 
we define the performance map for parameterized architectures. The selection of the best architecture is formulated as a multi-objective optimization problem. A family of architectures is then selected in the Pareto optimal sense. We also present the discovery and rediscovery of a particular optimal architecture by the preference function. 

\bigbreak
\noindent
\textbf{Multi-objective optimization and Pareto optimality}\\
\noindent
Consider, in general, a multi-objective optimization problem to minimize a vector valued map $J:X\to\Rr^N$: 
\begin{equation}\label{moo}
\min_{x \in X} J(x) := [J_1(x), J_2(x),\dots, J_N(x)]^T \ .
\end{equation}
In general, problem~\eqref{moo} does not necessarily admit a solution $x^*$ that minimizes all the objective functionals $\{J_i\}_{i \in \{1,2\dots, N\}}$.  A standard concept of solution for MOO problems is known as Pareto optimal solution. The definitions of Pareto optimal solution and Pareto front are as follows. 
\begin{definition}\label{def_pareto}
\begin{enumerate}
    \item For every $x^1,x^2 \in X$, we say that $x^1$ dominates $x^2$ if and only if $J_i(x^1) \leq J_i(x^2)$, for every $i \in \{1,2,\dots,N\}$, and there exists at least one $k$ such that $J_k(x^1)<J_k(x^2)$. 
    \item We say that $x\in X$ is a Pareto optimal solution if and only if there is no other element in $X$ that dominates $x$. We denote by $\Pp_J(X)$ the set of Pareto optimal solutions. 
    \item We call the Pareto front $\F_J(X)$ the image of the set of Pareto optimal solutions $\Pp_J(X)$ by the objective function $f$, that is, $\F_J(X) = \{J(x) \mid x \in \Pp_J(X) \}$. 
\end{enumerate}
\end{definition}

Note that the Pareto optimal and Pareto front can also be defined in the weak sense (see for instance ~\cite{censor1977pareto}), where strict inequalities are considered in Definition~\autoref{def_pareto}. 
\bigbreak
\noindent
\textbf{Selection of best architecture}\\
\noindent
Let us denote by $\Ss$ the parameterized space of different architectures. Note that $\Ss$ may vary depending on the context and tasks. 
We evaluate the performance of different architectures based on three criteria: (1) relative $\mathcal{L}_2$ error computed after applying the architecture to solve the real problem; (2) training time; and (3) number of parameters. 
These criteria can be modeled and quantified by a vector-valued performance map $f:\Ss \to \Rr^3$, where given an architecture $x\in\Ss$, $f_1(x)$ denotes the relative $\mathcal{L}_2$ error, $f_2(x)$ denotes the 
training time, and $f_3(x)$ is the number of parameters.  

As typically one wants to minimize all the aforementioned performance criteria, the selection of best architecture can then be formulated by a MOO problem such that: 
\begin{equation}\label{opt_ar}
\min_{x \in \Ss} f(x): = [f_1(x),f_2(x),f_3(x)]^T \ .
\end{equation}
We intend to identify the Pareto optimal architectures in the sense of optimization problem~\eqref{opt_ar}. 
Note that the MOO and Pareto optimal framework can be easily adapted when one wants to extend the domain $\Ss$, i.e., including more architectures, or extend the dimension of image set, i.e., considering more performance criteria. 
\bigbreak
\noindent
\textbf{Discovery of optimal architectures}\\
\noindent
We consider a preference function based discovery approach for finding the best architecture among the Pareto candidates. 
\begin{definition}
For a particular user, the discovery of optimal architecture refers to finding the best designed architecture that fits his preference. 
\end{definition}
In particular, the user's preference on the criteria is modeled by a preference function,  $p: \Rr^3 \to \Rr$, that maps the image set performance map $f$ to $\Rr$. Using this preference function, the selection of the best architecture among  Pareto optimal ones is reduced to a classical optimization problem of the form:
\begin{equation}
\min_{x \in \Pp_f(X)} p \circ f (x) \ .
\end{equation} 
An interesting property of the Pareto front is as follows.
\begin{lemma}[See~\cite{lee2024automatic}]\label{lemma_pare}
Assume $p$ is monotonically increasing. We have 
\begin{equation}
\argmin_{x \in X} p \circ {f}(x) = \argmin_{x \in \Pp_f(X)} p \circ {f}(x) \ .
\end{equation}
\end{lemma}
In light of~\Cref{lemma_pare}, as long as the user's preference function is monotonically increasing, it is sufficient to just evaluate the Pareto optimal architectures, in order to find the best one among all architectures. 

One natural choice for a preference function is the weighted sum function. However, to apply the weighted sum preference function, different performance maps must share the same image set. Thus, the performance data of the Pareto optimal architectures are first normalized to $\hat{f}$ using a re-scale function $r: \Rr^3 \to [0,1]^3$, such that $\hat{f}(x) = r(f(x))$ with 
\begin{equation}\label{re-scale}
\hat{f}_i(x) = r_i(f_i(x)) := \frac{f_i(x) - \underline{f_i}}{\overline{f_i} - \underline{f_i}}, \ \forall \ x \in \Pp_f(X) , \ \forall \ i \in \{1,2,3 \} \ ,
\end{equation}
where $\underline{f_i} = \min_{x \in \Pp_f(X)} f_i(x)$ and $\overline{f_i} = \max_{x \in \Pp_f(X)} f_i(x)$, and we assume $\underline{f_i}\neq \overline{f_i}$. 
In this case, the preference function has the form 
\begin{equation}\label{ws_pref}
p({f}) =  \sum_{\i=1}^3 \lambda_i r_i({f}_i) = \sum_{i=1}^3 \lambda_i \hat{f}_i  \ ,
\end{equation}
with $\sum \lambda_i = 1$. In this example, each weight $\lambda_i$ associated with $\hat{f}_i$ indicates how much the user cares about criterion $i$. 

\begin{remark}\label{rmk_ws}
The re-scale function $r_i$ is an increasing linear map on the performance data $f_i$ for every $i\in\{1,2,3 \}$. Applying this function on the performance data does not change the set of Pareto optimal solutions. Moreover, since the weighted sum function is also increasing with respect to the re-scaled performance data $\hat{f}$, \Cref{lemma_pare} holds.  
\end{remark}
\bigbreak
\noindent
\textbf{Rediscovery of architectures by linear programming}\\
\noindent
In this section, we explore the inverse direction of finding the best architecture. 
\begin{definition}
The rediscovery of architectures refers to identifying 
the specific preference function for a given Pareto optimal architecture  
under which it is optimal.
\end{definition}
We start by considering only the weighted sum preference function on the re-scaled data. 
Given a $x^* \in \Pp_f(X)$, 
identifying the weight $\lambda = (\lambda_1,\lambda_2,\lambda_3)$ 
can be formulated as a Linear Programming (LP) problem, of the form 
\begin{equation}
\begin{aligned}
&\min \lambda^T f(x^*) \\ 
& \text{ s. t. }
\left\{
\begin{aligned}
& \lambda^T f(x^*) \leq \lambda^T f(x) , \ \text{ for every } x \in \Pp_f(X) \text{ and } x \neq x^* \ , \\
& 0\leq \lambda_i \leq 1 \text{ and }  \sum_i\lambda_i=1 \ .
\end{aligned}
\right.
\end{aligned}
\end{equation}
This LP problem can be easily solved using modern LP solvers, e.g., CPLEX~\cite{cplex2009v12}. 

\subsection*{Iterative Pareto-optimal neighborhood sampling}
In this section, we present a simple and easily-adapted sampling based iterative procedure to explore the Pareto front. 
Recall that architectures are parametrized by a discrete design space $\Ss$,  and each $x\in \Ss$ represents a particular architecture.  
The performance of each architecture is evaluated by a multi-objective function \(f : \mathcal{S} \to \mathbb{R}^3\). 

By construction of the composite architectures, we have \(\mathcal{S} \subset \mathbb{N}^d\), where \(\mathbb{N} = \{0,1,2,\dots\}\) denotes the set of non-negative integers.  We equip \(\mathbb{N}^d\) with the componentwise partial order: for \(x,y \in \mathbb{N}^d\), we write \(x \leq y\) if and only if \(x_i \leq y_i\) for all \(i = 1,\dots,d\). 
Let \(\|\cdot\|\) denote the Euclidean norm on \(\mathbb{R}^d\). For \(x \in \mathcal{S}\) and \(r \ge 0\), we define the discrete ball
\[
    B(x,r) := \{\, y \in \mathcal{S} \mid \|y - x\| \le r \,\} \ .
\] 

Given a random seed $w$, the iterative procedure starts from a set of  fixed number $M$ randomly sampled parametrized architectures $\{x_1^\omega,x_2^\omega,\dots, x_M^\omega \} \subseteq \Ss$, denoted by $X^{\omega}_0$.  
Suppose that at iteration \(i\) we have a set of architectures
\(X_i^\omega\). 
We first extract the Pareto-optimal architectures \(\mathcal{P}_f(X_i^\omega)\)
within \(X_i^\omega\). Then the new set of architectures \(X_{i+1}^\omega\) is defined by
\begin{equation}
    X^w_{i+1} := \{ x \in \Ss \mid \exists \ y \in \Pp_f(X^w_i) \ \text{ s.t. } \ x \in B(y,r) \ \} \ .
\end{equation}
We repeat this procedure for a fixed number of $I$ iterations. In~\Cref{alg:pareto-sampling} we present a sketch of the sampling based algorithm. 

\begin{algorithm}[t]
\caption{Sampling-based exploration of the Pareto front}
\label{alg:pareto-sampling}
\begin{algorithmic}[1]
\Require Design space $\mathcal{S} \subset \mathbb{N}^d$, objective $f : \mathcal{S} \to \mathbb{R}^3$,
         radius $r > 0$, iterations $I \in \mathbb{N}$, sample size $M \in \mathbb{N}$, seed $\omega$
\State Initialize $X_0^\omega = \{x_1^\omega,\dots,x_M^\omega\} \subset \mathcal{S}$ using seed $\omega$
\For{$i = 0,1,\dots,I-1$}
    \State Evaluate $f(x)$ for all $x \in X_i^\omega$
    \State $\mathcal{P}_i \gets \mathcal{P}_f(X_i^\omega)$ \Comment{Pareto-optimal subset of $X_i^\omega$}
    \State $\widetilde{X}_{i+1} \gets \{ x \in \mathcal{S} \mid \exists\, y \in \mathcal{P}_i
            \text{ with } \|x - y\| \le r \}$
\EndFor
\State $\mathcal{P}_{I-1} \gets \mathcal{P}_f(X_{I-1}^\omega)$ \Comment{Pareto-optimal subset of $X_{I-1}^\omega$}
\State \Return $ \mathcal{P}_{I-1}$ as an approximation of the Pareto front
\end{algorithmic}
\end{algorithm}

\subsection*{Multi-objective evolutionary search via NSGA-II} 
In this section, we adapt the classical NSGA-II procedure to provide a multi-objective evolutionary baseline for our architecture space, for the particular glucose prediction task. 
Recall that architectures are parameterized by a discrete space $\Ss$. 
In this application, each architecture is specified by
\[
\begin{aligned}
&x=(x_1,x_2,x_3), \\
&x_1=[n,m,j,k]\in\{0,1,2\}^4\setminus\{(0,0,0,0)\}, \\ 
&x_2\in\{1,\dots,6\}, \\ 
&x_3\in\{16,32,64\}, 
\end{aligned}
\]
where the six values of $x_2$ correspond to the fixed block-type templates
(with LSTM always placed last), as described in the main text.
The performance of an architecture is evaluated by a bi-objective function
$f:\Ss\to \Rr^2$ to be minimized s.t.
\[
f(x)=\big(f_1(x),f_2(x)\big)
=\big(\text{relative }\ell_2\text{ error},\;\text{training time}\big). 
\]
Although the nominal number of configurations is $1440$, many are duplicates
because permuting absent blocks yields the same effective architecture;
after removing duplicates, the search space contains $  708 $ unique architectures.

To ensure that the evolutionary operators do not waste evaluations on duplicates,
we define a canonical representation.
Given $x=(x_1,x_2,x_3)$, we decode it into an expanded block sequence
$\sequ(x)$ by repeating each block-type according to its count in $x_1$
following the template specified by $x_2$,
and we define the canonical key
$\kappa(x):=(\sequ(x),x_3)$.
We precompute a dictionary that maps each canonical key $\kappa(x)$ to a unique
architecture identifier in the 708-element set.
Throughout NSGA-II, any newly generated candidate is first mapped to its unique
identifier via $\kappa(\cdot)$, then, candidates that have already been evaluated are rejected
and resampled (or mutated) to avoid wasting the evaluation budget. 

For two evaluated architectures $a,b\in\Ss$, we say that $a$ dominates $b$,
denoted $a\prec b$, if $f_m(a)\le f_m(b)$ for $m=1,2$ and strict inequality holds
for at least one objective. 
Given a population ${P}$, NSGA-II performs non-dominated sorting to partition
${P}$ into fronts ${F}_1,{F}_2,\dots$,
where ${F}_1$ is the non-dominated set of ${P}$ and subsequent fronts are
obtained by iteratively removing earlier ones. 
Each individual receives a rank $\rank(x)=k$ if $x\in {F}_k$. 
Within each front, we compute the crowding distance $\crowd(x)$ to promote diversity
along the Pareto front. 
In particular, for each objective, individuals are sorted by the objective value,
with boundary points are assigned infinite crowding distance, and interior points accumulate normalized neighbor gaps. 

Parents are selected from the entire current population by binary tournament selection, that is 
two individuals are drawn uniformly at random from ${P}$ and the one with smaller rank wins. Moreover, 
if ranks tie, the one with larger crowding distance wins. This yields an implicit selection pressure toward ${F}_1$ while preserving diversity. 
Offspring are generated using discrete operators on $x=(x_1,x_2,x_3)$: 
\begin{enumerate}
    \item uniform crossover: each component of the child is inherited from one of two parents with probability $1/2$; 
    \item mutation: with prescribed probabilities, we mutate (a) one count in $x_1$ by  $\pm 1$ step followed by clipping to $\{0,1,2\}$, (b) the template index $x_2$ by resampling in $\{1,\dots,6\}$, or (c) the hidden dimension $x_3$ by resampling in $\{16,32,64\}$.
\end{enumerate}
A repair step enforces feasibility, that is, if the mutation produces $x_1=(0,0,0,0)$, we randomly set one count to $1$.
Finally, the offspring is canonicalized via $\kappa(\cdot)$ and rejected if it duplicates an already evaluated architecture.

We emphasize that in our comparisons the total number of architecture evaluations is budgeted. In particular, 
let $B\in N$ be the evaluation budget and $P\in N$ the population size.
NSGA-II starts by evaluating an initial population of size $P$ (sampled uniformly from $\Ss$),
and then iterates until a total of $B$ distinct architectures have been evaluated.
At each step we track the non-dominated set among all evaluated architectures as the current approximation of the Pareto front. 
In~\Cref{alg:nsga2} we present a sketch of the NAGA-II based algorithm. 
\begin{algorithm}[t]
\caption{Budgeted NSGA-II for exploration of Pareto front}
\label{alg:nsga2}
\begin{algorithmic}[1]
\Require Unique architecture set $\Ss$, objectives $f:\Ss\to\Rr^2$,  
population size $P$, evaluation budget $B$, crossover prob. $p_c$, mutation prob. $p_m$, seed $\omega$
\State Initialize evaluated set $E\gets\emptyset$
\State Sample $P$ distinct architectures $\{x_1,\dots,x_P\}\subset\Ss$ using seed $\omega$
\State Evaluate $f(x_i)$ for $i=1,\dots,P$; set $E\gets\{x_1,\dots,x_P\}$; ${P}\gets\{x_1,\dots,x_P\}$
\While{$|E|<B$}
    \State Compute Pareto fronts $\{{F}_k\}$ of ${P}$ and ranks $\rank(\cdot)$
    \State Compute crowding distances $\crowd(\cdot)$ within each front
    \State ${Q}\gets\emptyset$ \Comment{offspring set}
    \While{$|{Q}|<P$ \textbf{and} $|E|+|{Q}|<B$}
        \State Select parents $p_1,p_2\in{P}$ by binary tournament
        \State With prob. $p_c$, generate child genotype by uniform crossover, otherwise copy one parent
        \State With prob. $p_m$, apply discrete mutation to $(x_1,x_2,x_3)$ and repair
        \State Canonicalize and map to a unique architecture $x_{\text{child}}\in\Ss$
        \If{$x_{\text{child}}\notin E\cup{Q}$}
            \State ${Q}\gets{Q}\cup\{x_{\text{child}}\}$
        \EndIf
    \EndWhile
    \State Evaluate $f(x)$ for all $x\in{Q}$; update $E\gets E\cup {Q}$
    \State ${R}\gets {P}\cup{Q}$ \Comment{elitist replacement}
    \State Non-dominated sort ${R}$ and compute crowding distances
    \State Form next population ${P}$ by filling fronts from ${F}_1$ upward, if a front cannot be fully included, keep individuals with largest crowding distance
\EndWhile
\State \Return $\mathcal{P}_f(E)$ as the Pareto-optimal set among all evaluated architectures
\end{algorithmic}
\end{algorithm}

\backmatter

\section*{Data Availability}
All processed data will be made available on Dropbox upon publication.

\section*{Code Availability}
We have demonstrated the glucose trajectory example on the dropbox. All the code are available on the GitHub repository \url{}.

\bibliography{sn-bibliography}

\section*{Acknowledgments}
We would like to acknowledge support by the U.S. Department of Energy (DOE), Office of Science, Advanced Scientific Computing Research (ASCR) program under the Scientific Discovery through Advanced Computing (SciDAC) Institute “LEADS: LEarning-Accelerated Domain Science,” Subcontract \#831126 under DE-AC05-76RL01830, DARPA-DIAL grant HR$00112490484$, and support from Shell, ExxonMobil, ABS, and Subsea 7.

\section*{Author Contributions}
\noindent
Q.C.: conceptualization, data curation, formal analysis, investigation, methodology, software, validation, visualization, writing--original draft, writing--review and editing.\\
S.L.: conceptualization, data curation, formal analysis, investigation, methodology, software, validation, visualization, writing--original draft, writing--review and editing.\\
A.J.V.: data curation, formal analysis, investigation, software, validation, writing--original draft.\\
J.D.: methodology, funding acquisition, project administration, supervision, writing---review and editing.\\
M.T.: methodology, funding acquisition, project administration, supervision, writing---review and editing.\\
G.E.K.: conceptualization, methodology, funding acquisition, project administration, resources, supervision, writing---review and editing.

\section*{Competing Interests}
Karniadakis provides technical advice on the direction of machine learning to PredictiveIQ and is a co-founder of PhinyxAI. The rest of the authors declare no competing interests.

\end{document}